\pdfoutput=1
\documentclass[11pt]{article}[]

\usepackage{acl}

\usepackage{times}
\usepackage{latexsym}
\usepackage{booktabs}
\usepackage{amssymb}
\usepackage{utfsym}
\usepackage[T1]{fontenc}
\usepackage[ruled,linesnumbered]{algorithm2e}
\usepackage[utf8]{inputenc}

\usepackage{microtype}

\usepackage{inconsolata}

\usepackage{graphicx}

\usepackage{amsthm}
\usepackage{amsmath}
\usepackage{amsfonts}
\usepackage{multirow}
\usepackage{booktabs}
\usepackage{cleveref}
\newcommand{\TheName}{CARFT}
\title{\TheName{}: Boosting LLM Reasoning via Contrastive Learning with Annotated Chain-of-Thought-based Reinforced Fine-Tuning}

\author{\textbf{Wenqiao Zhu\textsuperscript{1, 2, *}},
    \textbf{Ji Liu\textsuperscript{1, *}},
    \textbf{Rongjunchen Zhang\textsuperscript{1}},
    \textbf{Haipang Wu\textsuperscript{1}},
    \textbf{Yulun Zhang\textsuperscript{2}}
    \\
  \textsuperscript{1}HiThink Research,
  \textsuperscript{2}Shanghai Jiao Tong University \\    
}

\begin{document}
\maketitle
\def\thefootnote{*}\footnotetext{Corresponding authors: zhuwnq@outlook.com and jiliuwork@gmail.com}
\begin{abstract}

Reasoning capability plays a significantly critical role in the the broad applications of Large Language Models (LLMs). To enhance the reasoning performance of LLMs, diverse Reinforcement Learning (RL)-based fine-tuning approaches have been proposed to address the limited generalization capability of LLMs trained solely via Supervised Fine-Tuning (SFT). Despite their effectiveness, two major limitations hinder the advancement of LLMs. First, vanilla RL-based approaches ignore annotated Chain-of-Thought (CoT) and incorporate unstable reasoning path sampling, which typically results in model collapse, unstable training process, and suboptimal performance. Second, existing SFT approaches generally overemphasize the annotated CoT, potentially leading to performance degradation due to insufficient exploitation of potential CoT. In this paper, we propose a Contrastive learning with annotated CoT-based Reinforced Fine-Tuning approach, i.e., \TheName{}, to enhance the reasoning performance of LLMs while addressing the aforementioned limitations. Specifically, we propose learning a representation for each CoT. Based on this representation, we design novel contrastive signals to guide the fine-tuning process. Our approach not only fully exploits the available annotated CoT but also stabilizes the fine-tuning procedure by incorporating an additional unsupervised learning signal. We conduct comprehensive experiments and in-depth analysis with three baseline approaches, two foundation models, and two datasets to demonstrate significant advantages of \TheName{} in terms of robustness, performance (up to 10.15\%), and efficiency (up to 30.62\%).
Code is available at https://github.com/WNQzhu/CARFT.

\end{abstract}

\section{Introduction}

The reasoning capability of Large Language Models (LLMs) stands as a critical component, driving an extensive array of potential applications, which span mathematical problem \citep{wang2024mathcoder,WizardMath23}, financial analysis \citep{yang2023fingpt,zhang2023fingptrag}, and medical applications \citep{Clinical22}, etc. The advent of reasoning LLMs, e.g., OpenAI o1 \citep{jaech2024openai}, OpenAI o3 \citep{openai-o3}, Llama-Nemotron \citep{bercovich2025llama}, Claude 3.7 \citep{Claude}, and DeepSeek R1 \citep{r1}, has significantly heightened the interest in exploring the reasoning capabilities of LLMs across both academic and industrial sectors. Additionally, given the straightforward verification of answers, the task of solving mathematical problems has emerged as a pivotal domain in the study of LLM reasoning capacities.

One of the conventional strategies for augmenting the reasoning capabilities of LLMs is Supervised Fine-Tuning (SFT). SFT entails fine-tuning LLMs with training samples that incorporate annotated Chain-of-Thought (CoT) \citep{10.5555/3600270.3602070}. In a training dataset $\mathcal{D}_\text{train}$, each training sample is structured as a tuple $( \mathbf{x},  \mathbf{c}, \mathbf{y} )$, where $\mathbf{x}$ represents the input question, $\mathbf{c}$ represents the annotated CoT, and $\mathbf{y}$ denotes the correct ground truth answer. CoT $\mathbf{c}$ in the training sample is generally written or labeled by experienced experts or high-end LLMs, which is highly valuable for the fine-tuning of LLMs.

SFT-based reasoning enhancement approaches only exploit a single annotated CoT for each question within the training dataset. However, multiple CoTs \citep{zhang-etal-2023-interpretable} exist for each question. Hence, conducting SFT with only a single annotated CoT in the training dataset may limit the generalization capability of LLMs. 

\begin{table}[t]
\begin{center}
\begin{tabular}{ccc}
\toprule
Method & A-CoT & SG-CoT \\
\midrule
SFT  &  \usym{2714}  & \usym{2718} \\ 
PPO-like (e.g., ReFT) &  \usym{2718} & \usym{2714} \\
\TheName{} & \usym{2714} & \usym{2714} \\
\bottomrule
\end{tabular}
\end{center}
\caption{An overview of whether methods employ Annotated-CoT (A-CoT) or Self-Generated CoT (SG-CoT).}
\label{tbl:use-cot}
\end{table}

To address the limitations of SFT-based methods, Reinforcement Learning (RL)-based fine-tuning approaches emerge \citep{luong2024reft, grpo, dr.grpo}. A prominent and state-of-the-art RL-based fine-tuning approach is ReFT \citep{luong2024reft}, which incorporates online RL approach, i.e., Proximal Policy Optimization (PPO) \citep{Schulman2017ProximalPO}, to dynamically sample CoTs at each training step. This mechanism enables ReFT to leverage multiple CoTs, thereby improving the generalization capability of LLMs.

\begin{figure*}[t]
\centering
\includegraphics[width=\linewidth]{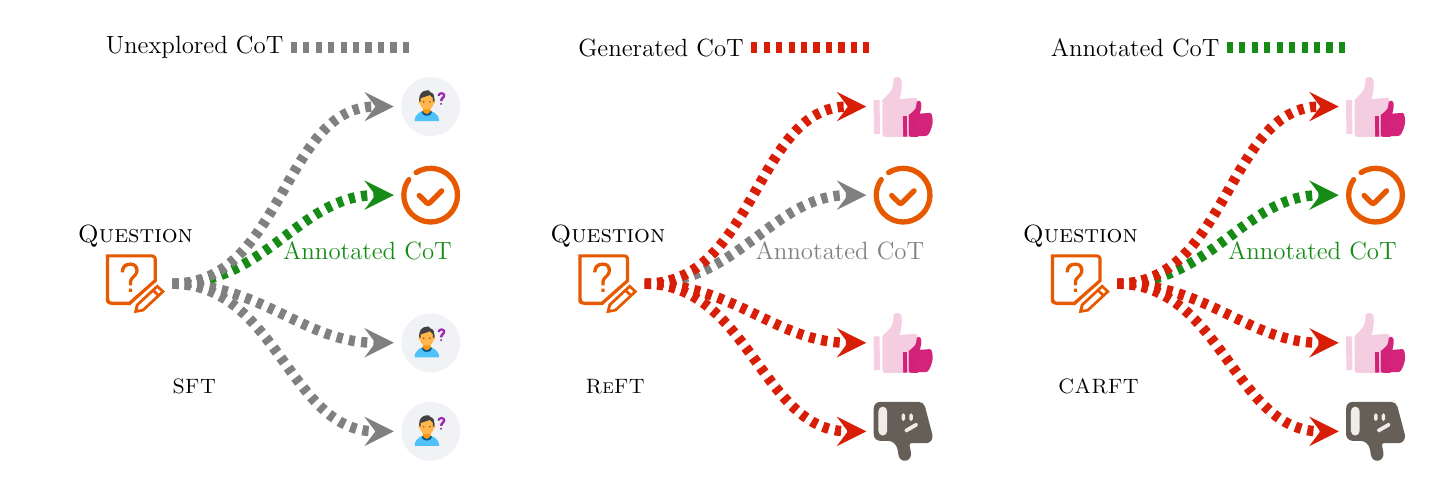}
\caption{Comparison between SFT, ReFT, and \TheName{} on the exploration of CoT.}
\label{fig:use-cot}
\end{figure*}

Despite their effectiveness, two major limitations still exist with the existing RL-based fine-tuning approaches, which hinders the advancement of LLMs. First, existing RL-based approaches solely rely on on-policy sampled CoTs without considering the highly valued annotated CoTs while enhancing reasoning performance. Due to the reward hacking problem \citep{10.5555/3600270.3600957}, such sampled CoTs may not be valid or correct, potentially degrading model performance. Second, existing approaches suffer from unstable training. The inherent exploratory nature of RL can lead to model collapse, i.e., a phenomenon where the behavior of LLMs significantly deteriorates during training. This instability can severely impact the performance of LLMs and result in undesirable outputs.

To address these challenges, we propose a novel Contrastive learning with Annotated CoT-based Reinforced Fine-Tuning approach, i.e., \TheName{}, which effectively leverages the valuable annotated CoTs in the training dataset while sampling other potential CoTs so as to achieve superb performance. \TheName{} begins with learning a unified representation for each CoT, encompassing both high-quality annotated CoTs and on-policy sampled CoTs. Based on this representation, we design contrastive signals to improve both the reasoning performance and the stability of the fine-tuning process. Specifically, we propose exploiting a masked loss, e.g., InfoNCE \citep{infonce}, to utilize the unified representation to generate the contrastive signal. This contrastive signal serves as a guiding mechanism for the on-policy CoT sampling process, helping to stabilize the fine-tuning of LLMs while maximizing the utilization of information from the annotated CoT. Table \ref{tbl:use-cot} and Figure \ref{fig:use-cot} illustrate the working characteristic of \TheName{}. In addition, we propose embedding-enhanced partial reward to further improve the performance. The key contributions of this paper are summarized as follows:
\begin{itemize}
\item We propose a novel contrastive learning-based framework with an original contrastive signal construction method that fully exploits annotated CoTs to improve both the performance and the stability in the fine-tuning of LLMs. 
\item We design an embedding-enhanced partial reward so as to 
 further improve the stability in the reinforced fine-tuning process and to achieve superb performance of LLMs.
\item We conduct extensive experiments and thorough ablation studies to demonstrate the effectiveness of \TheName{} compared with three baseline approaches, two foundation models, and two datasets. Extensive experimental results demonstrate that \TheName{} significantly outperformances baselines in terms of effectiveness (up to 10.15\%) and robustness.
\end{itemize}

\section{Related Work}

\paragraph{Reinforcement Learning (RL)-based LLM Reasoning.} Recent years have witnessed widespread application in Natural Language Processing (NLP), particularly in the domains of preference optimization \citep{Stiennon@LearningFromHF@2020, rafailov2024direct, azar24aipo, zhu-etal-2025-sgdpo} and reasoning \citep{luong2024reft, grpo, dapo, dr.grpo}. These methods typically follow a standard three-stage pipeline: (1) SFT, (2) reward modeling, and (3) RL-based optimization. A key distinction among these approaches lies in how the reward signal is obtained. In preference optimization, reward models are learned from human feedback, while in mathematical reasoning tasks, rule-based methods are typically exploited to construct reward signals, as ground-truth answers can be explicitly verified. Within the context of preference optimization, Direct Preference Optimization (DPO) \citep{rafailov2024direct} has emerged as an effective algorithm that avoids the need for explicit reward model training. However, due to its offline nature \citep{feng2024@dpo_grad_flow}, DPO may struggle to explore diverse CoTs \citep{luong2024reft}. As a result, on-policy approaches, e.g., GRPO \cite{grpo}, DAPO \citep{dapo}, Dr.GRPO \citep{dr.grpo}, and ReFT \cite{luong2024reft}, are generally employed to better explore such diversity in reasoning.

On-policy approaches utilize multiple rollouts to estimate the Generalized Advantage Estimation (GAE). DAPO and Dr.GRPO are both improved variants of GRPO. Specifically, DAPO is designed for long-CoT scenarios and introduces four key techniques: higher clipping, dynamic sampling, token-level policy gradient loss, and overlong reward shaping. On the other hand, Dr.GRPO improves upon GRPO by eliminating the bias present in the original method. While these approaches are effective, they come with the trade-off of increased computational complexity. In contrast, ReFT \citep{luong2024reft} utilizes only a single on-policy sample per step, making it significantly computationally efficient.

Despite their strengths, these approaches rely solely on on-policy sampling, ignoring potentially valuable annotated CoTs already present in the training data. Moreover, model collapse occurs frequently within the reinforced fine-tuning process with the existing approaches.

\paragraph{Contrastive Learning.} Contrastive learning has shown strong effectiveness in diverse fields such as multimodal pretraining \citep{radford2021clip}, recommendation systems \citep{kgcl2022}, graph embedding \citep{10.1145/3511808.3557704}, and report generation \citep{zhou-wang-2024-divide}. Theoretically, the contrastive loss can be decomposed into two components: an alignment term and a uniformity term \citep{10.5555/3524938.3525859}. The alignment term reduces the distance between embeddings of positive pairs, while the uniformity term encourages embeddings of negative pairs to be uniformly dispersed in the representation space. Inspired by this framework, we propose in this paper to utilize contrastive feedback to guide the online generation of CoT reasoning.
\section{Method}

In this section, we first present the preliminary of reinforced LLM fine-tuning. Then, we detail \TheName{}, including the contrastive learning-based framework with an original contrastive signal construction method and an embedding
enhanced partial reward method.

\begin{figure*}[t]
\centering
\includegraphics[width=\linewidth]{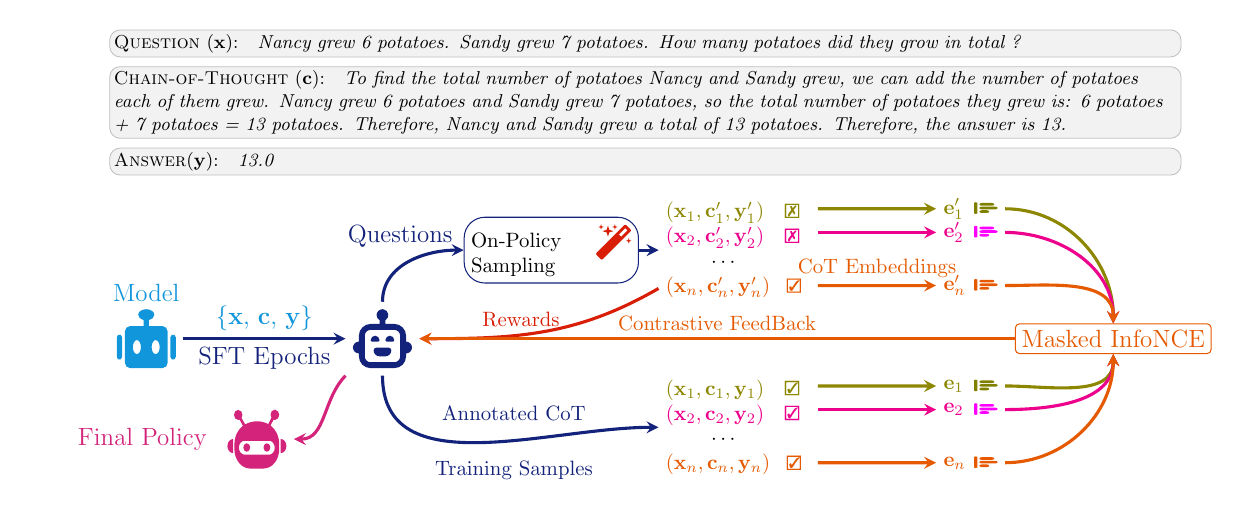}
\caption{The framework \TheName{} is composed of two sequential stages: (i) supervised fine-tuning (SFT), followed by (ii) contrastive feedback.}
\label{fig:framework}
\end{figure*}

\subsection{Preliminary of Reinforced Fine-Tuning}

Reinforced fine-tuning incorporates feedback signals derived from either a learned reward model or predefined rules to guide the training of LLMs. Given an input prompt $ x $ and the corresponding response $ y $ produced by a LLM, the objective is to maximize the expected cumulative reward, which is formally formulated as:
\begin{align}
\max_{\pi_\theta}  & \mathbb{E}_{x\sim \mathcal{D}, y\sim\pi_\theta(y|x)}\left[r(x,y)\right] \nonumber \\ 
&- \beta \mathbb{D}_{\text{KL}}\left[\pi_\theta(y|x)||\pi_\text{ref}(y|x)\right],
\end{align}
where $ r(x, y) $ denotes the reward function, $ \theta $ represents the parameters of the LLM, $ \pi_\theta $ refers to the learnable target policy, $ \pi_\text{ref} $ corresponds to a reference policy, i.e., typically the initially pre-trained LLM, exploited to stabilize training, and $ \beta $ denotes the coefficient of KL-divergence, which encourages the updated policy to stay close to the original distribution.

A commonly used algorithm in this setting is PPO \citep{Schulman2017ProximalPO}, which employs GAE \citep{Schulman2015HighDimensionalCC} for stable gradient updates. In particular, the advantage at time step $ t $ is calculated as:
\begin{equation}
\label{eq:advantage}
\hat{A}_t = \sum_{l=0}^{L-t}(\gamma \lambda)^l\sigma_{t+l}, 
\end{equation}
where $L$ represents the maximum length of token sequence, $ \sigma_{t'} $ is the Temporal Difference (TD) residual at timestep $ t^\prime $, $ \lambda \in (0, 1] $ serves as the GAE discount factor, and $ \gamma \in [0, 1] $ controls the discounting of TD residuals. $ \sigma_{t'} $ is formulated as:
\begin{align}
\sigma_{t^\prime} = -V_\phi(s_{t^\prime}) + r_\text{total}(s_{t^\prime}, a_{t^\prime}, s_{t^{\prime} + 1}) + \gamma V_\phi(s_{t^\prime}) \nonumber,
\end{align}
where $ s_{t^\prime} $ denotes the state at time step $ t^\prime$, and $ a_{t^\prime} $ represents the action at time step $ t^\prime$, $ \gamma $ is similar as defined in Formula \ref{eq:advantage}, $r_\text{total}(\cdot)$ calculates the total reward, and $ V_\phi(\cdot) $ estimates the state value from a given state with $\phi$ referring to a vector of policy parameters. Actions correspond to individual tokens selected from the vocabulary.

The total reward at the token level includes both the external reward signal and an internal regularization based on the Kullback–Leibler (KL) divergence between the current and reference policies, as defined as follows:
\begin{align}
r_\text{total}(s_{t^\prime}, a_{t^\prime}, s_{t^{\prime} + 1}) & = r(s_{t^\prime}, a_{t^\prime}) \nonumber \\ & + \beta \mathbb{D}_{\text{KL}}\left[\pi_\theta(\cdot|s_{t^\prime}) ||
\pi_\text{ref}(\cdot|s_{t^\prime})\right] \nonumber,
\end{align}
where $\pi_\theta$ represents the sampling actions policy with policy parameters $\phi$, and $\pi_\text{ref}$ corresponds to the sampling actions with a reference policy. Given GAE $\hat{A}_t$ and state value $V_\phi(s_t)$, we can estimate the reward as defined in Formula \ref{eq:reward}:
\begin{equation}
\label{eq:reward}
    \hat{R}_t = \hat{A}_t + V_\phi(s_t).
\end{equation}
Under the PPO framework, the policy and value loss functions are separately formulated to ensure a stable and effective fine-tuning process. The policy loss is defined in Formula \ref{eq:policy}.
\begin{align}
\label{eq:policy}
\mathcal{L}_\text{policy}(\theta) &= -\mathbb{E}\left[
\min\left(
\frac{\pi_\theta(a_t|s_t)}{\pi^\text{old}_\theta(a_t|s_t)}\hat{A}_t, \right.\right.\nonumber \\
&\left.\left.\text{clip}\left(\frac{\pi_\theta(a_t|s_t)}{\pi^\text{old}_\theta(a_t|s_t)}, 1 - \epsilon, 1 + \epsilon\right)\hat{A}_t
\right)
\right],
\end{align}
where $\pi^\text{old}_\theta$ corresponds to the sampling process before update, $\epsilon$ represents a hyperparameter controlling the clipping range and preventing excessively large policy updates, and clip$\left(\frac{\pi_\theta(a_t|s_t)}{\pi^\text{old}_\theta(a_t|s_t)}, 1 - \epsilon, 1 + \epsilon\right)\hat{A}_t$ modifies the surrogate objective by clipping the probability ratio, which removes the incentive for moving $\frac{\pi_\theta(a_t|s_t)}{\pi^\text{old}_\theta(a_t|s_t)}$ outside of the interval [1 - $\epsilon$, 1 + $\epsilon$] (see details in \citep{Schulman2017ProximalPO}).
Then, the value loss is defined as:
\begin{align}
\mathcal{L}_\text{value} &(\phi) = \frac{1}{2}\mathbb{E}\left[
\max\left(
\Vert V_\phi(s_t) - \hat{R}_t \Vert^2, \right.\right.\nonumber \\
&\left\Vert\left.\left.\text{clip}\left(
\hat{R}_t  - V_\phi(s_t), 
\hat{A}_t - \epsilon,
\hat{A}_t + \epsilon
\right)\right\Vert^2
\right)
\right], \nonumber
\end{align}
where $ \epsilon $ is similar as defined in \ref{eq:policy}. Finally, the overall reinforcement learning loss combines both the policy loss and the value loss as defined in Formula \ref{eq:rlloss}:
\begin{equation}
\label{eq:rlloss}
\mathcal{L}_{\text{RL}} = \mathcal{L}_\text{policy} + \alpha \mathcal{L}_\text{value},
\end{equation}
where $ \alpha $ balances the relative importance of the policy and value losses within the reinforced fine-tuning process.

\subsection{\TheName{}}

While existing approaches either overemphasizes the annotated CoT (for SFT) or face challenges in achieving stable reinforced fine-tuning while ignoring annotated CoT (for existing RL-based approaches), we propose a novel contrastive learning-based approach, i.e., \TheName{}, to properly levarage the annotated CoTs so as to address this issue. In this section, we first present the overall workflow of \TheName{}. Then, we explain the CoT embeddings. Afterward, we propose masked contrastive signal construction approach in \TheName{}. Finally, we explain a novel embedding enhanced partial reward method.

\subsubsection{Workflow}

As shown in Figure \ref{fig:framework}, the overall workflow of \TheName{} consists of two sequential stages: the SFT stage and the reinforced fine-tuning stage.

\paragraph{SFT} We assume that each training sample in the training dataset is a triplet $(\mathbf{x}, \mathbf{c}, \mathbf{y})$, where $\mathbf{x}$ denotes the input question, $\mathbf{c}$ represents the annotated CoT, and $\mathbf{y}$ is the ground-truth answer. We carry out SFT with a few epochs to improve the instruction-following ability of the LLM.

\paragraph{Contrastive Feedback} Let $\mathbf{c}_1$ and $\mathbf{c}_2$ denote two distinct CoTs corresponding to the same input question $\mathbf{x}_1$, derived either from training examples or on-policy sampling, i.e., $(\mathbf{x}_1, \mathbf{c}_1, \mathbf{y}_1)$ and $(\mathbf{x}_1, \mathbf{c}_2, \mathbf{y}_1)$. We assume that $\mathbf{y}_1$ is the valid answer to $\mathbf{x}_1$. Since $\mathbf{c}_1$ and $\mathbf{c}_2$ pertain to the same input $\mathbf{x}_1$, we posit the existence of a conditional distribution $p_1(\mathbf{c} \mid \mathbf{h}_1)$ such that both $\mathbf{c}_1 \sim p_1(\mathbf{c} \mid \mathbf{h}_1)$ and $\mathbf{c}_2 \sim p_1(\mathbf{c} \mid \mathbf{h}_1)$, where $\mathbf{h}_1$ denotes a latent variable associated with $\mathbf{x}_1$. Given that $\mathbf{c}_1$ and $\mathbf{c}_2$ are sampled from the same distribution, there should exist a similarity metric $m(\cdot, \cdot)$ under which the distance between $\mathbf{c}_1$ and $\mathbf{c}_2$ is smaller than the distance between $\mathbf{c}_1$ and any $\mathbf{c}_i$ drawn from a different distribution $p_{i, i \neq 1}(\mathbf{c} \mid \mathbf{h}_i)$, with a high probability:
\begin{equation*}
m(\mathbf{c}_1, \mathbf{c}_2) \leq m(\mathbf{c}_1, \mathbf{c}_i), \quad \text{for } \mathbf{c}_i \sim p_{i, i \neq 1}(\mathbf{c} \mid \mathbf{h}_i).
\end{equation*}
This insight provides two key advantages when incorporated as an unsupervised signal. First, it enables us to exploit the annotated CoTs in the training data in the reinforced fine-tuning process of LLMs. Second, it offers a guiding signal for CoT generation, helping to stabilize the reinforced fine-tuning process and to mitigate the risk of model collapse.

\subsubsection{Chain-of-Thought Embeddings}
\label{subsubsec:cotEmb}

Given a CoT $\mathbf{c}$ of length $L$, represented as:
\begin{equation*}
\mathbf{c} = \left[a_1, a_2, \cdots, a_L\right],
\end{equation*}
we denote the corresponding token embeddings and state values as:
\begin{equation*}
\mathbf{H} = \left[H_1, H_2, \cdots, H_L\right]
\end{equation*}
and 
\begin{equation*}
\mathbf{V}_\phi = \left[V_\phi(1), V_\phi(2), \cdots, V_\phi(L)\right],
\end{equation*}
respectively.

To obtain a compact representation of the entire CoT, we compute a weighted sum of the token embeddings using the softmax-normalized state values, as defined in Formula \ref{eq:representation}.
\begin{equation}
\label{eq:representation}
\mathbf{e} = \sum \text{Softmax}(\mathbf{V}_\phi) \odot \mathbf{H},
\end{equation}
where $\odot$ denotes element-wise multiplication between the state values and the corresponding embedding vectors.

In practice, in order to reduce memory consumption, we first project each embedding $H_i$ into a lower-dimensional space using a simple single-layer MultiLayer Perceptron (MLP), denoted by $\text{proj}(\cdot)$. The projected embeddings are then exploited in place of the original ones:
\begin{equation*}
\mathbf{H} = \left[\text{proj}(H_1), \text{proj}(H_2), \cdots, \text{proj}(H_L)\right].
\end{equation*}

\subsubsection{Masked Contrastive Signal Construction}
\label{subsubsec:signal}

In this section, we design two types of contrastive signals for reinforced fine-tuning, i.e., positive and negative. We denote the signal related to CoT that results in a correct answer by positive signal, and that results in a wrong answer by negative signal.

\begin{algorithm*}
  \SetKwInOut{Input}{Input}
  \SetKwInOut{Output}{Output}
  
  \Input{Tuples of (\textit{question, CoT, answer}): $\mathcal{D}_\text{train} = \{(\mathbf{x}, \mathbf{c}, \mathbf{y}\}$, Number of RL steps: $T$, Number of updates per RL step:$U$, Initial policy: $\mathbf{\pi}_\theta^0$. }
  \Output{Final Policy: $\mathbf{\pi}_\theta$ }
  %%% \emph{special treatment of the first line}\;
  \For{$i\leftarrow 1$ \KwTo $T$}{
  $\mathbf{x}, \mathbf{c}, \mathbf{y} \sim \mathcal{D}_\text{train}$  \qquad\tcp{Sample training data from $\mathcal{D}_\text{train}$}
  $\hat{\mathbf{c}} \sim \pi_\theta$ \qquad\tcp{On-policy CoT sampling}
  
  $\hat{\mathbf{y}} \leftarrow \textsc{Extract}(\hat{\mathbf{c}})$ \qquad\tcp{Extract answer}

  $\mathbf{e}^\text{annotated} \leftarrow \mathbf{c}$, \quad $\hat{\mathbf{e}}^\text{rollout} \leftarrow \hat{\mathbf{c}}$ \qquad\tcp{Construct CoT Embeddings}

  Compute $\sigma_t, \hat{A}_t, \hat{R}_t, \mathcal{M}_1$
  
   \For{$i\leftarrow 1$ \KwTo $U$}{
    $\theta, \phi \leftarrow $\textsc{Optimization\_step}($\mathcal{L}$)  \qquad\tcp{Equation \ref{eq:r3ft:opt}}
   }
  }
  \Return{$\pi_\theta$}
\LinesNumberedHidden
\caption{\TheName{} with Positive Signal}\label{algo:r3ft:pos}
\end{algorithm*}

\paragraph{Positive Signal} Given a batch of training samples $\{\mathbf{x}_i, \mathbf{c}_i^\text{annotated}, \mathbf{y}_i\}_1^B$ with $B$ presenting the batch size, 
we conduct LLM self-generation to generate a batch of rollout CoTs, i.e., $\{\mathbf{x}_i, \mathbf{c}_i^\text{rollout}, \mathbf{y}_i\}_1^B$. By employing the CoT embedding module, we could get embeddings of the annotated CoTs $\{\mathbf{e}^\text{annotated}_i\}_1^B$ and rollout CoT $\{\mathbf{e}^\text{rollout}_i\}_1^B$ exploiting the approach presented in Section \ref{subsubsec:cotEmb}.
We construct a contrastive feedback with InfoNCE \citep{infonce} to provide the positive contrastive signal as defined in Formula \ref{eq:positive}.
We describe the \TheName{} framework with positive signal in Algorithm~\ref{algo:r3ft:pos}. For each pair $(\mathbf{c}, \hat{\mathbf{c}})$, where $\mathbf{c}$ denotes the annotated CoT and $\hat{\mathbf{c}}$ represents the on-policy sampled CoT, we construct the corresponding embeddings for each CoT, resulting in $\mathbf{e}^\text{annotated}$ and $\hat{\mathbf{e}}^\text{rollout}$, respectively.
Which is then utilized to guide the fine-tuning steps.

\begin{equation}
\label{eq:positive}
\mathcal{L}_{c_1} =  \sum_{i=1}^B-\log\frac{\exp(\langle \mathbf{e}^\text{annotated}_i,\mathbf{e}^\text{rollout}_i \rangle/\tau)\odot \mathcal{M}_1}{\sum_{j=1}^B\exp(\langle \mathbf{e}^\text{annotated}_i,\mathbf{e}^\text{rollout}_j\rangle/\tau)},
\end{equation}
where $\mathcal{M}_1$ represents a binary mask, in which each element takes the value 1 if the corresponding CoT leads to a correct answer, and 0 otherwise. The notation $\langle \cdot, \cdot \rangle$ denotes the inner product.

\paragraph{Negative Signal} We devise a scheme to utilize the signal within the negative CoT as well. We denote the annotated CoTs and the associated negative CoTs by $\mathbf{c'}_i^{\text{annotated}}$ and $\mathbf{c'}_i^{\text{rollout}}$, respectively. Initially, we calculate the Longest Common Subsequence (LCS) of $\mathbf{c'}_i^{\text{annotated}}$ and $\mathbf{c'}_i^{\text{rollout}}$. Subsequently, based on the LCS and the parts of the sequence that exclude the LCS, we construct four embeddings, denoted $\mathbf{e}_{i, \text{LCS}}^{\text{annotated}}$, $\mathbf{e}_{i, \text{exc}}^{\text{annotated}}$, $\mathbf{e}_{i, \text{LCS}}^{\text{rollout}}$, and $\mathbf{e}_{i, \text{exc}}^{\text{rollout}}$, respectively. Then, the negative contrastive signal is formulated in Formula \ref{eq:negative}.
\begin{equation}
\label{eq:negative}
\mathcal{L}_{c_2} =  \sum_{i = 1}^B -\log\frac{\exp(\langle \mathbf{e}_{i,\text{LCS}}^{\text{rollout}}, \mathbf{e}_{i,\text{exc}}^{\text{annotated}} \rangle/\tau)\odot \mathcal{M}_2}{\sum_{j = 1}^B\exp(\langle \mathbf{e}_{i,\text{LCS}}^{\text{rollout}}, \mathbf{e}_{j,\text{exc}}^{\text{rollout}}\rangle/\tau)},
\end{equation}
where $\mathcal{M}_2$ represents a binary mask, in which each element takes the value 1 if the corresponding CoT leads to a wrong answer, and 0 otherwise.
See details in Appendix.

\paragraph{Optimization} We optimize the following reinforcement learning loss to learn the policy:
\begin{equation}
    \mathcal{L} = \mathcal{L}_\text{RL} + c \left\{\mathcal{L}_{c_1} \,\text{or} \: \mathcal{L}_{c_2} \right\}
    \label{eq:r3ft:opt}
\end{equation}
where $ c $ balances the relative importance of the PPO and contrastive losses during the reinforced fine-tuning process (see  detailed algorithms in Appendix).

\subsubsection{Embedding-enhanced Partial Reward}
\label{subsubsec:overfitting}

In order to further improve the stability and the performance of the contrastive signal, we propose an embedding-enhanced partial reward method.

ReFT \citep{luong2024reft} assigns a partial reward $ r(x, y) = 0.1 $ to the CoT when it is a negative CoT, from which a numerical answer can be extracted. Unlike the partial reward in ReFT, we introduce a fine-grained partial reward by leveraging our unified CoT Embedding, which provides a tool to measure CoT similarity.
\begin{equation}
r(x,y) = \langle \mathbf{e}^\text{annotated},\mathbf{e}^\text{rollout}\rangle \times 0.1 + 0.2.
\end{equation}
The inner product $\langle\mathbf{e}^\text{annotated}, \mathbf{e}^\text{rollout}\rangle$ ranges from $-1$ to $1$, leading to a partial reward range of $[0.1, 0.3]$. When the CoTs are dissimilar, the inner product approaches $-1$, resulting in a reward close to $0.1$; when they are similar, the reward approaches $0.3$. This strategy encourages well-behaved CoT generation. By assigning differentiated rewards to negative CoTs, the embedding-enhanced partial reward method further improve the stability of the reinforced fine-tuning process and the final performance of LLMs.

\begin{table}[t]
\begin{center}
\begin{tabular}{ccc}
\toprule
Method & SVAMP & GSM8K \\
\midrule
\#Train Samples & 3076 & 7465 \\
\#Test Samples  & 1000 & 1319 \\
\bottomrule
\end{tabular}
\end{center}
\caption{Statics of the train and test datasets.}
\label{tbl:stat}
\end{table}

\section{Experiments}

In this section, we present the experimental results. First, we present our experiment setup. Then, we demonstrate the evaluation of \TheName{} compared with SFT, ReFT, and Dr.GRPO. Afterward, we present an ablation study. 

\subsection{Experimental Setup}

%%% Datasets

We conduct experiments on two publicly available datasets: SVAMP \citep{patel-etal-2021-nlp} and GSM8K \citep{cobbe2021gsm8k}. Table \ref{tbl:stat} presents the key statistics of SVAMP  and GSM8K. For the reasoning process, we leverage the CoT annotations from \citep{luong2024reft}, which were generated based on few-shot prompting \citep{10.5555/3600270.3602070,10.5555/3618408.3618843} with GPT-3.5-turbo \citep{GPT_3_5_turbo}. Our experiments are conducted based on two foundation models: CodeLlama-7B \citep{codellama} and Qwen2.5-7B-Instruct \citep{qwen2_5}. We evaluate \TheName{} in comparison with three baseline approaches: SFT, ReFT \citep{luong2024reft}, and Dr.GRPO \citep{dr.grpo}. ReFT is a state-of-the-art RL approach for LLM fine-tuning. As an advanced extension of GRPO \citep{grpo}, Dr.GRPO demonstrates excellent performance on R1-like \citep{r1} tasks. See setup details in Appendix.

\subsection{Evaluation of \TheName{}}

As illustrated in Table \ref{tbl:avg_acc}, \TheName{} significantly outperforms SFT and ReFT across different models (up to 10.15\% on average). With the SVAMP dataset, \TheName{} yields substantial accuracy enhancements compared with SFT. Precisely, the accuracy escalates from 62.3\% to 64.8\% and from 86.9\% to 88.0\%, with absolute increments of 2.5\% and 1.1\% for CodeLlama and Qwen2.5-Instruct, respectively. 
Moreover, with the GSM8K dataset, \TheName{} showcases remarkable improvements as well. The accuracy climbs from 43.82\% to 50.95\% and from 80.67\% to 84.31\%, corresponding to absolute boosts of 7.13\% and 3.64\%  for CodeLlama and Qwen2.5-Instruct, respectively.  

Table \ref{tbl:avg_acc} further reveals that ReFT can outperform the SFT baseline once the training process stabilizes. Nevertheless, the performance of ReFT remains inferior to \TheName{}. In addition, the experimental results demonstrate that ReFT is plagued by the model collapse issue, which significantly undermines its effectiveness. Furthermore, we find that when model collapse occurs, the performance of ReFT lags far behind that of SFT (up to 14.56\%) and \TheName{} (up to 18.2\%). 

\begin{table*}[t]
\begin{center}
\begin{tabular}{ccccc}
\toprule
Method & Size & SVAMP & GSM8K & Average \\
\midrule
CodeLlama + SFT & 7B& 62.3\% & 43.82 \% & 53.06\% \\
CodeLlama + ReFT \citep{luong2024reft} &7B& \underline{62.5}\% & \underline{50.27\%} & \underline{56.39\%} \\
CodeLlama + \TheName{} & 7B& \textbf{64.8\%}& \textbf{50.95\%} & \textbf{57.88\%} \\
Qwen2.5-Instruct + SFT & 7B & \underline{86.9\%} & \underline{80.67\%}& \underline{83.79\%}\\
Qwen2.5-Instruct + ReFT \citep{luong2024reft} & 7B & 85.9\% &66.11\% & 76.01\% \\
Qwen2.5-Instruct + \TheName{} & 7B & \textbf{88.0\%} & \textbf{84.31\%} & \textbf{86.16\%}\\
\bottomrule
\end{tabular}
\end{center}
\caption{Evaluation Accuracy of Various Methods on the SVAMP and GSM8K Datasets.}
\label{tbl:avg_acc}
\end{table*}

\begin{figure}[t]
\centering
\includegraphics[width=\linewidth]{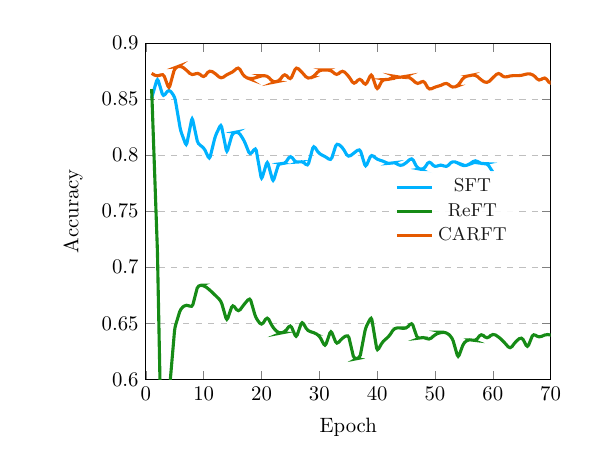}
\caption{Accuracy curves of various methods on SVAMP dataset and Qwen2.5-7B-Instruct backbone.}
\label{fig:acc:qwen:svamp}
\end{figure}

Furthermore, Figures \ref{fig:acc:qwen:svamp} depicts the accuracy curves of different approaches on the SVAMP dataset, with Qwen2.5-7B-Instruct serving as the backbone model. These results indicate that ReFT undergoes model collapse after undergoing fine-tuning for just one epoch. In addition, we find that when using Qwen2.5-7B-Instruct, SFT is prone to an unstable tuning process, as evidenced by the decline in its accuracy as the fine-tuning process advances. In contrast, \TheName{} exhibits remarkable stability and superb performance throughout the entire training process. This outstanding performance can be attributed to our contrastive feedback mechanism as presented in Section \ref{subsubsec:signal}, which offers reference signals for the generation of CoTs. 

\begin{figure}[t]
\centering
\includegraphics[width=\linewidth]{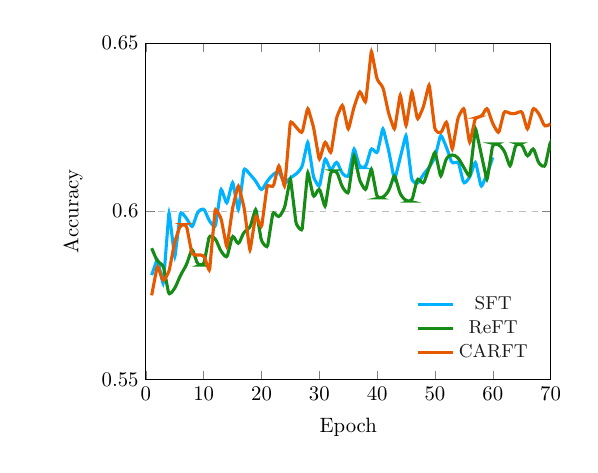}
\caption{Accuracy curves of various methods on SVAMP dataset and CodeLlama-7B backbone.}
\label{fig:acc:codellama:svamp}
\end{figure}

As demonstrated in Figure \ref{fig:acc:codellama:svamp}, \TheName{} consistently outperforms ReFT during the fine-tuning process and converges rapidly, swiftly reaching peak accuracy values. Nevertheless, the figure also suggests that \TheName{} is potentially vulnerable to unstable fine-tuning process. See additional experimental results in Appendix. 

As shown in Table~\ref{tbl:cmp-dr.grpo}, \TheName{} with embedding-enhanced partial reward enabled significantly outperforms all baseline approaches in terms of both accuracy (up to 0.5\% compared with Dr.GRPO and 1.7\% compared with ReFT) and efficiency (up to 30.62\% compared with Dr.GRPO). Interestingly, Dr.GRPO also surpasses ReFT in terms of performance metrics (1.2\%), which is accompanied with a considerable increase in computational time. Specifically, Dr.GRPO relies on significant computing resources due to the generation of a larger number of CoTs.
 
\begin{table}[t]
\begin{center}
\begin{tabular}{ccc}
\toprule
Method &  Accuracy & Time Cost(hours) \\
\midrule
ReFT &62.5\% &  \textbf{14.12} \\
Dr.GRPO &  \underline{63.7\%}& 24.49 \\
\TheName{} & \textbf{64.2\%}& \underline{16.99}\\
\bottomrule
\end{tabular}
\end{center}
\caption{Evaluation Accuracy of Various Methods on the SVAMP Datasets, based on CodeLlama-7B.}
\label{tbl:cmp-dr.grpo}
\end{table}

\subsection{Ablation Study}

\begin{figure}[t]
\centering
\includegraphics[width=\linewidth]{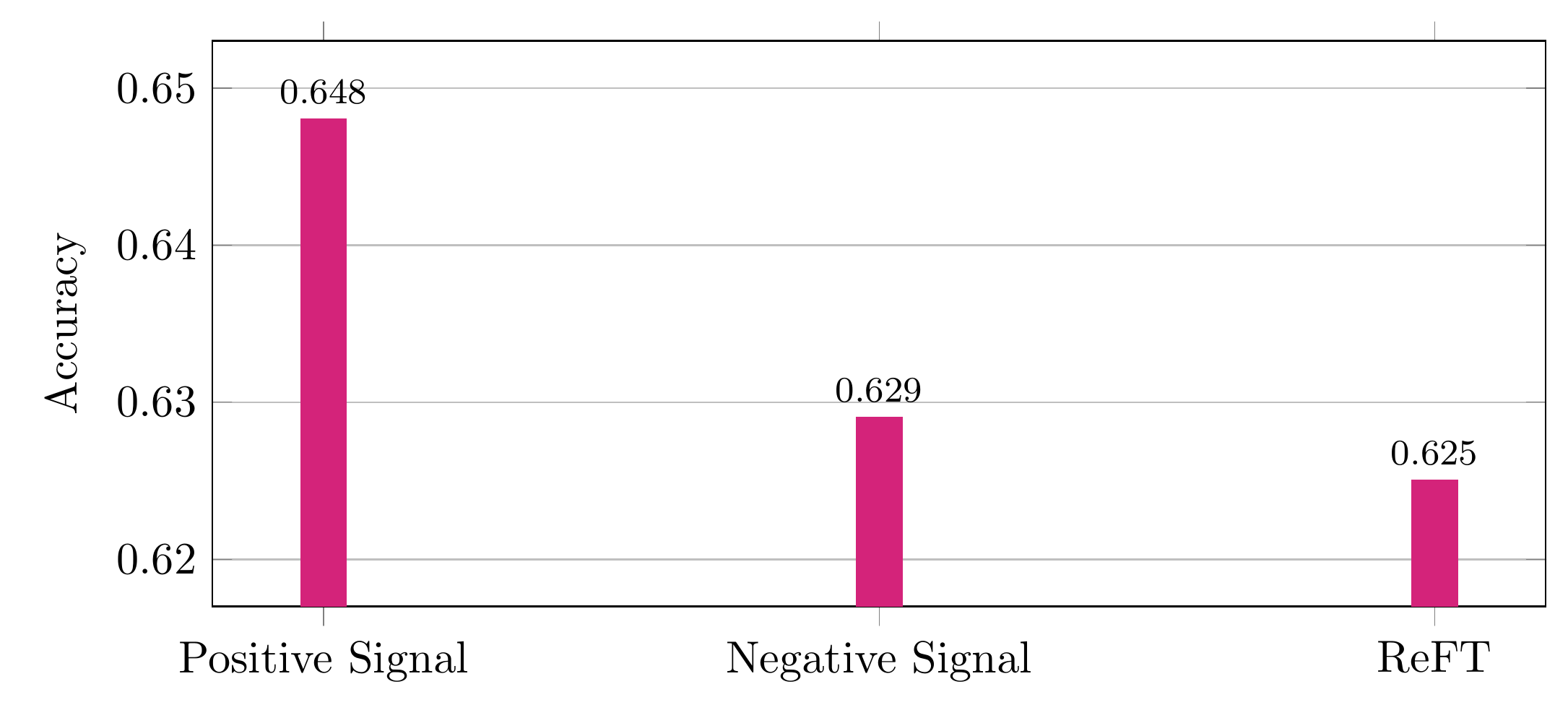}
\caption{Accuracy of \TheName{} with positive signal and negative signal, based on the SVAMP dataset and with the CodeLlama-7B as the backbone model.}
\vspace{-4mm}
\label{fig:acc-bar:pos-and-neg}
\end{figure}

\begin{figure}[t]
\centering
\includegraphics[width=\linewidth]{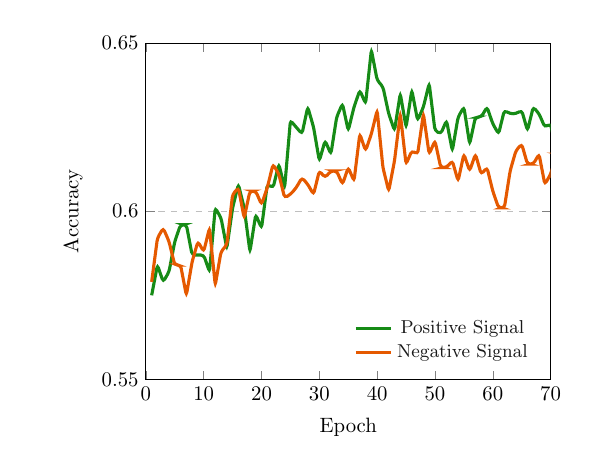}
\caption{Accuracy Curve of \TheName{} with positive signal and negative signal, based on the SVAMP dataset and with the CodeLlama-7B as the backbone model.}
\label{fig:acc-curve:pos-and-neg}
\end{figure}

\paragraph{Positive Signal versus Negative Signal}
We conduct an ablation study to show the impact of positive and negative contrastive signals. As shown in Figure \ref{fig:acc-bar:pos-and-neg}, \TheName{} outperformances ReFT with both positive (2.30\% higher) and negative contrastive (0.4\% higher) signals. Notably, the positive signal demonstrates a more pronounced performance gain (1.9\% higher) compared to its negative counterpart. Due to its excellent performance, we employ the positive signal in our experiments. 

\begin{figure}[t]
\centering
\includegraphics[width=\linewidth]{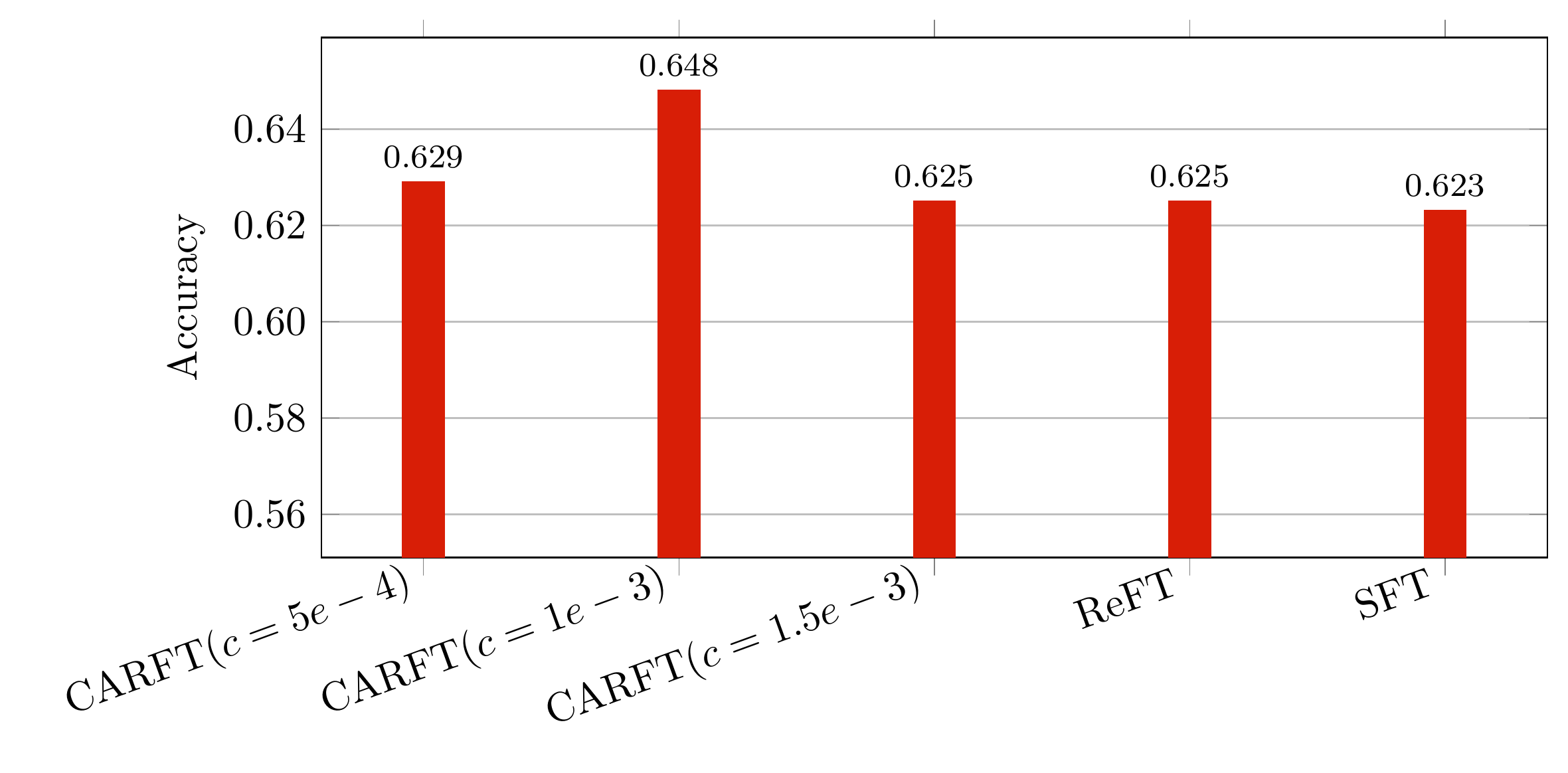}
\vspace{-7mm}
\caption{Accuracy of \TheName{} with different $c$, based on the SVAMP dataset and with the CodeLlama-7B as the backbone model.}
\label{fig:acc-bar:codellama:cl_coef}
\vspace{-4mm}
\end{figure}

\paragraph{Robustness}
To assess the robustness of the proposed method, we perform an ablation study on the contrastive loss coefficient $c$. As illustrated in Figure \ref{fig:acc-bar:codellama:cl_coef}, by systematically varying the value of $c$ within the range from $5\times10^{-4}$ to $1.5\times10^{-3}$, we observe that \TheName{} consistently outperforms the SFT and ReFT baseline across all tested values. This consistent superiority in performance strongly validates the robustness of \TheName{} and demonstrates its resilience to changes in the contrastive loss coefficient. 

\paragraph{Stability}

\begin{figure}[t]
\centering
\includegraphics[width=\linewidth]{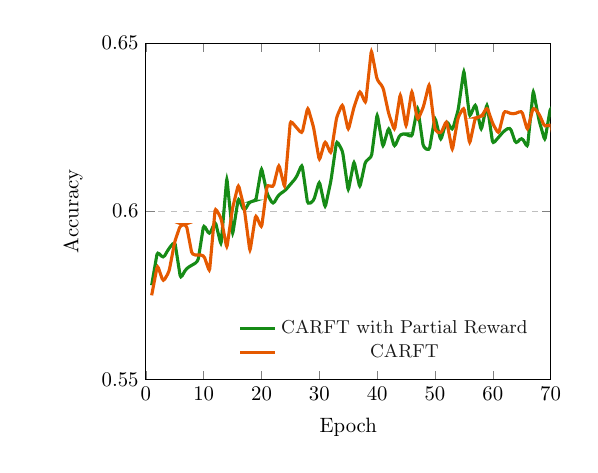}
\caption{Accuracy Curve of \TheName{} Partial Rewards, based on the SVAMP dataset and with the CodeLlama-7B as the backbone model.}
\label{fig:acc-curve:codellama:aug-and-partial-reward}
\end{figure}

To further enhance the stability of the reinforced fine-tuning process, we propose an embedding-enhanced partial reward method, as described in Section \ref{subsubsec:overfitting}. As shown in Figure \ref{fig:acc-curve:codellama:aug-and-partial-reward}, this approach effectively improves training stability. The tuning process of \TheName{} using the embedding-enhanced partial reward method achieves a final accuracy that is 0.5\% higher than that of the baseline without the method. Moreover, \TheName{} with this enhancement exhibits a more stable accuracy improvement curve. \TheName{} with embedding-enhanced partial reward enabled achieves a peak accuracy of 64.2\%, which also corresponds to significant improvements over ReFT, with gains of up to 1.7\%.


\subsection{Complexity}
Simpler methods like SFT require significantly less computational overhead as they do not involve the rollout process used in RL-based methods. However, the performance of SFT is lower compared to RL-based approaches. Our proposed CARFT method achieves the highest performance among all the considered methods.

Let $ N $ denote the number of parameters in the LLM, and $ L $ represent the length of the on-policy sampled CoT. The computational complexity of a single forward pass in ReFT is $ O((L + 1) \cdot N) $, while that of SFT is $ O(N) $. Since Dr.GRPO requires multiple rollouts, let $ G $ be the number of rollouts. Accordingly, the computational complexity of Dr.GRPO becomes $ O((G \cdot L + 1) \cdot N) $.  As for CARFT, which incorporates a contrastive signal, let $ M $ represent the size of the projector and $ d $ denote the hidden dimension of the LLM. The computational complexity of CARFT is then given by $ O((L + 1) \cdot N + M \cdot d \cdot L) $, where $ M \cdot d $ is significantly smaller than $ N $.

The space complexity of SFT is O(N). In contrast, ReFT incurs an additional overhead of O(Ld) due to caching the sequence of on-policy CoT trajectories, resulting in a total space complexity of $ O(N + L\cdot d) $.  
Similarly, Dr.GRPO  has a space complexity of $ O(N + G \cdot L \cdot d). $
For \TheName{}, if we denote the projected hidden size as $d_2$, its space complexity becomes $O(N + L\cdot d + M + L \cdot d_2)$. Here, $ M $ is significantly smaller than $ N $, and $d_2$ (the compressed hidden dimension) is much smaller than the original hidden size $ d $.

\section{Conclusions}
In this paper, we propose a novel contrastive learning-based framework with annotated CoTs, i.e., \TheName{}, to enhance the reasoning capabilities of LLMs. We propose generating contrastive signals from both positive and negative CoTs while incorporating annotated CoTs. In order to further improve the stability of the reinforced fine-tuning process, we propose a novel embedding-enhanced partial reward method. Extensive experimental results demonstrate significant advantages of \TheName{} in terms of performance (up to 10.15\%) and efficiency (up to 30.62\%). In addition, \TheName{} corresponds to better stability during reinforced fine-tuning compared with existing approaches.

\section*{Limitations}

\TheName{} requires additional computational overhead to compute the embeddings of CoTs, to achieve excellent. As a result, it consumes a longer computational time compared to the ReFT and SFT. However, \TheName{} needs less computational time than Dr.GRPO as \TheName{} needs less on-policy sampled CoTs. In addition, \TheName{} is designed to exploit a centralized annotated CoT dataset. The annotated datasets may be stored in multiple data centers or devices, which may hinder the application of \TheName{} with decentralized data. In addition, we have restricted the context to fewer than 1024 tokens in this work. We plan to explore improving reasoning in long-context scenarios \citep{zhu2024psc}. In addition, while \TheName{} exploits centralized data, we will investigate decentralized datasets, e.g., federated learning \cite{Liu2024Fisher,liu2024efficient,jia2024efficient,liu2024aedfl,liu2024fedasmu,Che2023Federated,liu2023distributed,liu2022multi,Zhang2022FedDUAP,zhou2022efficient,jia2025efficient,liu2025efficient,chen2025trustworthy,liu2024enhancing} and distributed machine learning \cite{liu2023heterps}.
\bibliography{emnlp_2025}

%%%\newpage
\cleardoublepage
\appendix
\section{Appendix}
\label{sec:appendix}

\subsection{Experimental Settings}
All experiments are conducted on an ensemble of 8 H100-80GB GPUs. Given that reinforced fine-tuning for reasoning tasks is inherently time-intensive, we utilize FlashAttention \citep{dao2022flashattention,dao2023flashattention2} and DeepSpeed Zero stage 3 \citep{10.5555/3433701.3433727,10.1145/3394486.3406703} to expedite the fine-tuning process. These technologies enable us to scale up the batch size, thereby enhancing computational efficiency.  
Additionally, we utilize the HuggingFace Alignment Handbook \citep{Tunstall_The_Alignment_Handbook} and the TRL library \citep{vonwerra2022trl} as methodological guides to streamline the fine-tuning implementation. 

To ensure consistency and comparability across experiments, we adopt a structured hyperparameter configuration strategy. During the warmup phase, we initialize training with a batch size of 64 and a learning rate of 1e-5. This learning rate is then adjusted to 3e-7 during the reinforcement fine-tuning stage to stabilize the optimization process. We maintain a batch size of 64 for all models on the SVAMP dataset. On GSM8K, we tailor the batch size to each model's computational characteristics: 64 for Qwen2.5-7B-Instruct and 96 for CodeLlama-7B, balancing memory efficiency and training throughput. In reinforcement learning components, we set the KL divergence coefficient to 0.05 to regulate policy updates and employ a temperature parameter ($\tau$) of 0.2 and $c=1e-3$ in the contrastive learning loss to control embedding similarity. We set the dimension of the projected embedding to 64. For PPO optimization, we configure $\lambda = 1$, $\gamma = 0.95$, $\alpha = 5$, $\epsilon = 0.2$, and $U = 2$.  For Dr.GRPO, to ensure a fair comparison, we set the parameter $G$ in Dr.GRPO to 2, which matches the maximum number of CoTs in \TheName{} at each step. 
We set the reward $ r(x,y) $ to 1 if the answer is correct, and 0 otherwise. We also adopt a partial reward scheme \citep{le2022coderl}, setting the reward to 0.1 in cases where a numerical answer can be extracted but is incorrect.

Training epoch limits are determined based on empirical convergence behavior. For the SFT baseline, we cap training at 60 epochs due to its tendency to be unstable; beyond this point, additional epochs yield diminishing returns. To ensure a fair comparison across methods, for ReFT, \TheName{}, and Dr.GRPO, we fine-tune the base model for 4 epochs and select the best checkpoint for reinforced fine-tuning. All of these approaches are then trained for 70 epochs, allowing sufficient iterations for convergence while maintaining experimental rigor.
\subsection{More Experiments}

\begin{table}[t]
\begin{center}
\begin{tabular}{ccc}
\toprule
Batch Size & Accuracy  \\
\midrule
64 & \textbf{51.48\%} \\
96 & 50.95\% \\
\bottomrule
\end{tabular}
\end{center}
\caption{ \TheName{} with different batch size, based on CodeLlama-7B model and GSM8K dataset.}
\label{app:tbl:batch-size}
\end{table}

\paragraph{Batch Size} In our experiments, we utilized FlashAttention \citep{dao2022flashattention,dao2023flashattention2} and DeepSpeed Zero stage 3 \citep{10.5555/3433701.3433727,10.1145/3394486.3406703} to accelerate the fine-tuning process with a large batch size. To evaluate how batch size affects model performance, we conducted a systematic ablation study. As shown in Table~\ref{app:tbl:batch-size}, increasing the batch size can degrade the performance of large language models (LLMs). Specifically, enlarging the batch size from 64 to 96 led to a drop in accuracy from 51.48\% to 50.95\%. 

This suggests that reducing the batch size may be a viable strategy for achieving better performance. It is worth emphasizing that all experiments were carried out with consistent batch size configurations to ensure a fair and valid comparison.

\begin{table}[t]
\begin{center}
\begin{tabular}{ccc}
\toprule
Model & Model & Accuracy \\
\midrule
Qwen2.5-Instruct-14B  & SFT  & 86.7\%\\ 
Qwen2.5-Instruct-14B  & \TheName{} & 88.9\%\\
\bottomrule
\end{tabular}
\end{center}
\caption{Accuracy curves of various methods on SVAMP dataset and Qwen2.5-14B-Instruct backbone.}
\label{app:tbl:larger-model}
\end{table}
\paragraph{Larger Model}
In addition to the two 7B models, we conducted experiments on Qwen 2.5 14B, as well. The models were fine-tuned for 100 epochs, and the results are summarized in Table \ref{app:tbl:larger-model}. These results highlight the effectiveness and strong generalization of \TheName{} across diverse model scales and datasets.

\paragraph{Explains of the Accuracy Curve}
\begin{figure}[t]
\centering
\includegraphics[width=\linewidth]{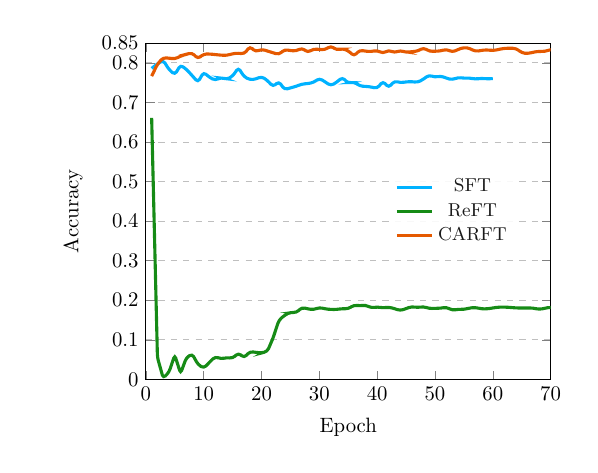}
\caption{Accuracy curves of various methods on GSM8K dataset and Qwen2.5-7B-Instruct backbone.}
\label{app:fig:acc:qwen:gsm8k}
\end{figure}

Figure \ref{app:fig:acc:qwen:gsm8k} indicates that ReFT also suffers from model collapse, which yields poor results. \TheName{} shows strong stability across the whole fine-tuning process and outperforms both SFT and ReFT significantly.

\begin{figure}[t]
\centering
\includegraphics[width=\linewidth]{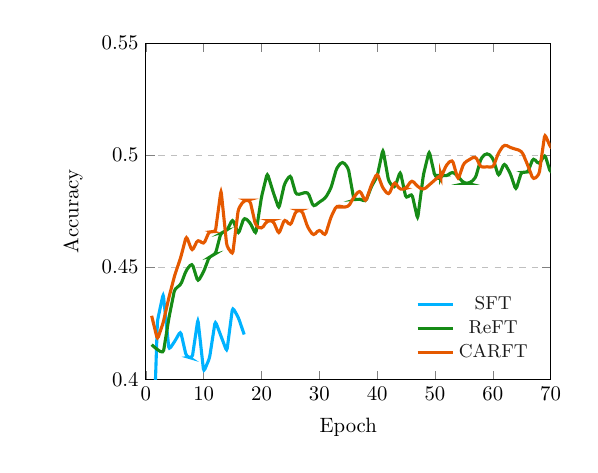}
\caption{Accuracy curves of various methods on GSM8k dataset and CodeLlama-7B backbone.}
\label{app:fig:acc-curve:codellma-gsm8k}
\end{figure}

Figure~\ref{app:fig:acc-curve:codellma-gsm8k} presents the accuracy of various methods on the GSM8K dataset using the CodeLlama-7B model as the backbone. We observed that further training did not lead to performance improvements in the SFT (Supervised Fine-Tuning) phase, so we terminated the training early. The figure also demonstrates that \TheName{} outperforms ReFT with a higher convergency accuracy.

\begin{figure}[t]
\centering
\includegraphics[width=\linewidth]{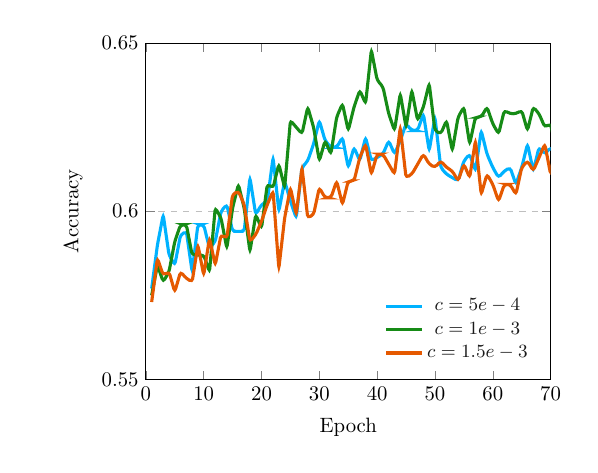}
\caption{Accuracy Curve of \TheName{} with different $c$, based on the SVAMP dataset and with the CodeLlama-7B as the backbone model.}
\label{app:fig:acc-curve:codellama:cl-coef}
\end{figure}

Figure \ref{app:fig:acc-curve:codellama:cl-coef} illustrates the accuracy curves of \TheName{} under different values of parameter $c$. It can be observed that \TheName{} attains the optimal performance when $c = 1e-3$.

\begin{figure}[t]
\centering
\includegraphics[width=\linewidth]{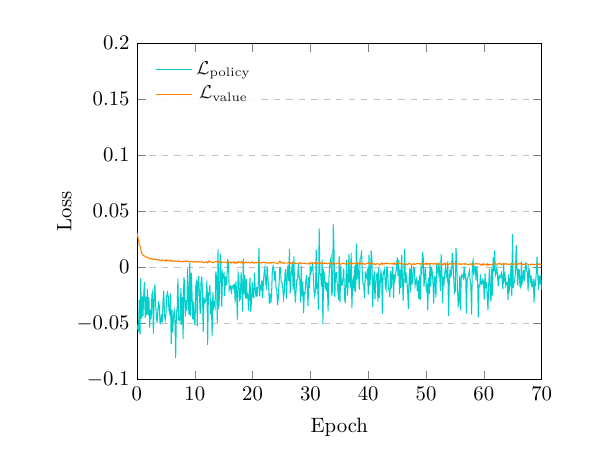}
\caption{RL loss curve for \TheName{} with the GSM8K dataset and CodeLlama-7B serving as the backbone model.}
\label{app:fig:loss:codellama:r3ft}
\end{figure}

\begin{figure}[t]
\centering
\includegraphics[width=\linewidth]{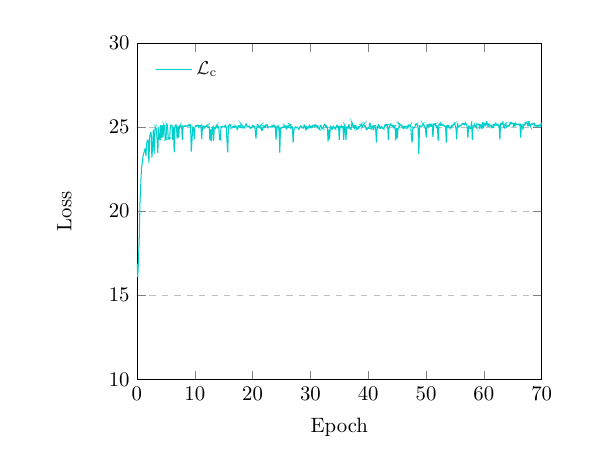}
\caption{Constrastive loss curve for \TheName{} with the GSM8K dataset and CodeLlama-7B serving as the backbone model.}
\label{app:fig:cl-loss:codellama:r3ft}
\end{figure}

\begin{figure}[t]
\centering
\includegraphics[width=\linewidth]{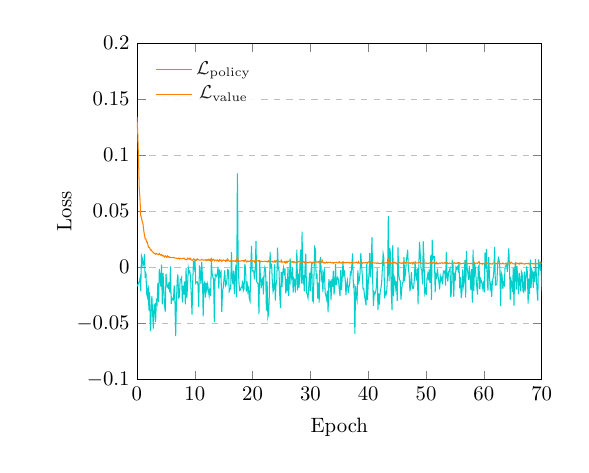}
\caption{RL loss curve for ReFT with the GSM8K dataset and CodeLlama-7B serving as the backbone model.}
\label{app:fig:loss:codellama:reft}
\end{figure}

\begin{figure}[t]
\centering
\includegraphics[width=\linewidth]{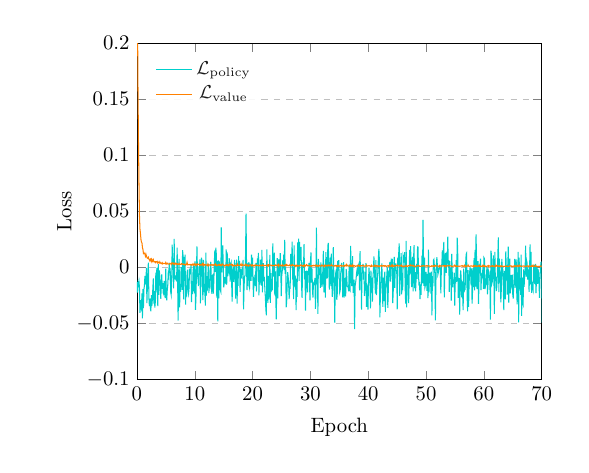}
\caption{RL loss curve for \TheName{} with the GSM8K dataset and Qwen2.5-7B-Instruct serving as the backbone model.}
\label{app:fig:loss:qwen_2_5:r3ft}
\end{figure}

\begin{figure}[t]
\centering
\includegraphics[width=\linewidth]{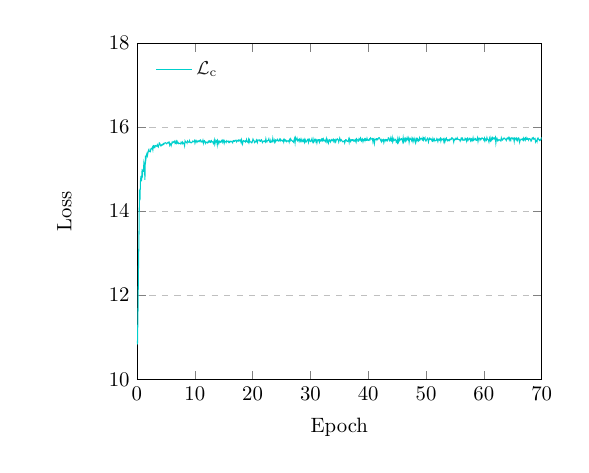}
\caption{Contrastive loss curve for \TheName{} with the GSM8K dataset and Qwen2.5-7B-Instruct serving as the backbone model.}
\label{app:fig:cl-loss:qwen_2_5:r3ft}
\end{figure}

\begin{figure}[t]
\centering
\includegraphics[width=\linewidth]{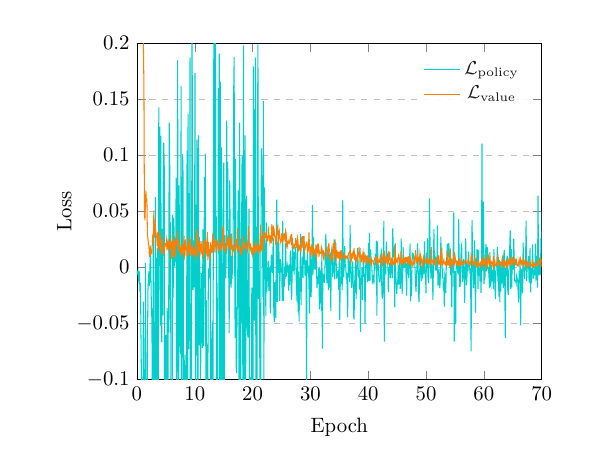}
\caption{RL loss curve for ReFT with the GSM8K dataset and Qwen2.5-7B-Instruct serving as the backbone model.}
\label{app:fig:loss:qwen_2_5:reft}
\end{figure}

\paragraph{Loss}
We present the RL (Reinforcement Learning) and contrastive learning loss curves for \TheName{} and ReFT models in Figures \ref{app:fig:loss:codellama:r3ft}, \ref{app:fig:cl-loss:codellama:r3ft}, \ref{app:fig:loss:codellama:reft}, \ref{app:fig:loss:qwen_2_5:r3ft}, \ref{app:fig:cl-loss:qwen_2_5:r3ft}, and \ref{app:fig:loss:qwen_2_5:reft}. These results are based on the GSM8K dataset and utilize the CodeLlama-7B and Qwen2.5-7B-Instruct base models. 
We make the following observations:  
(1) As shown in Figure \ref{app:fig:loss:codellama:r3ft} and Figure \ref{app:fig:loss:codellama:reft}, when using CodeLlama-7B as the base model, \TheName{} and ReFT exhibit similar loss curves.  
(2) In contrast, when Qwen2.5-7B-Instruct is used as the base model, the loss curves of \TheName{} and ReFT differ significantly. In particular, ReFT displays a fluctuating pattern, suggesting instability during fine-tuning.  
(3) Furthermore, as seen in Figure \ref{app:fig:cl-loss:codellama:r3ft} and Figure \ref{app:fig:cl-loss:qwen_2_5:r3ft}, the contrastive loss varies across models, indicating differences in the CoT embedding spaces learned by each model.

%%% qwen 2.5

\begin{algorithm*}
  \SetKwInOut{Input}{Input}
  \SetKwInOut{Output}{Output}  
  \Input{Tuples of (\textit{question, CoT, answer}): $\mathcal{D}_\text{train} = \{(\mathbf{x}, \mathbf{c}, \mathbf{y}\}$, Number of RL steps: $T$, Number of updates per RL step:$U$, Initial policy: $\mathbf{\pi}_\theta^0$. }
  \Output{Final Policy: $\mathbf{\pi}_\theta$ }
  %%% \emph{special treatment of the first line}\;
  \For{$i\leftarrow 1$ \KwTo $T$}{
  $\mathbf{x}, \mathbf{c}, \mathbf{y} \sim \mathcal{D}_\text{train}$  \qquad\tcp{Sample training data from $\mathcal{D}_\text{train}$}
  $\hat{\mathbf{c}} \sim \pi_\theta$ \qquad\tcp{On-policy CoT sampling}
  $\hat{\mathbf{y}} \leftarrow \textsc{Extract}(\hat{\mathbf{c}})$ \qquad\tcp{Extract answer}

    $\mathbf{e}^\text{annotated}_\text{LCS}, \mathbf{e}^\text{annotated}_\text{exc} \leftarrow \mathbf{c}$, \quad    $\hat{\mathbf{e}}^\text{rollout}_\text{LCS}, \hat{\mathbf{e}}^\text{rollout}_\text{exc} \leftarrow \hat{\mathbf{c}}$ \qquad\tcp{Construct CoT Embeddings}

  Compute $\sigma_t, \hat{A}_t, \hat{R}_t, \mathcal{M}_2$
  
   \For{$i\leftarrow 1$ \KwTo $U$}{
    $\theta, \phi \leftarrow $\textsc{Optimization\_step}($\mathcal{L}$)  \qquad\tcp{Equation \ref{eq:r3ft:opt}}
   }
  }
  \Return{$\pi_{\theta}$}
\LinesNumberedHidden
\caption{\TheName{} with Negative Signal}\label{algo:r3ft:neg}
\end{algorithm*}

% \subsection{\TheName{} with Positive Signal}
\subsection{\TheName{} with Negative Signal}
We describe the \TheName{} framework with negative signal in Algorithm~\ref{algo:r3ft:neg}. For each pair $(\mathbf{c}, \hat{\mathbf{c}})$, where $\mathbf{c}$ denotes the annotated CoT and $\hat{\mathbf{c}}$ represents the on-policy sampled CoT, we first compute the longest common subsequence (LCS) between the two sequences. Using the LCS tokens, we construct corresponding LCS embeddings for both sequences, resulting in $\mathbf{e}^\text{annotated}_\text{LCS}$ and $\hat{\mathbf{e}}^\text{rollout}_\text{LCS}$, respectively. The remaining tokens—that is, those not included in the LCS—are used to generate two additional embeddings: $\mathbf{e}^\text{annotated}_\text{exc}$ and $\hat{\mathbf{e}}^\text{rollout}_\text{exc}$.

The masked InfoNCE loss then leverages $\mathbf{e}^\text{rollout}_\text{LCS}$, $\mathbf{e}^\text{annotated}_\text{exc}$, and $\hat{\mathbf{e}}^\text{rollout}_\text{exc}$ to provide feedback for training.

\subsubsection{Illustration}
We provide an example to illustrate how negative signals are constructed.
Let  $$ A = [a_1, a_2, \dots, a_n] $$  
denote the annotated CoT that leads to a correct solution.  
Let  
$$ B = [b_1, b_2, \dots, b_m] $$  
represent the on-policy sampled CoT that results in an incorrect solution.

\paragraph{Step 1: Compute Longest Common Subsequence (LCS)}
We first compute the LCS of $ A $ and $ B $, denoted as:
$$ C = [c_1, c_2, \dots, c_k] $$

\paragraph{Step 2: Extract Sub-sequences}
Next, we compute the embeddings of the following components:
\begin{itemize}
\item[-] $ C $: the common subsequence,
\item[-] $ A \setminus C $: parts of the correct CoT not in $ C $,
\item[-] $ B \setminus C $: parts of the incorrect CoT not in $ C $.
\end{itemize}

\paragraph{Step 3: Apply Contrastive Signal}
We minimize the contrastive loss to generate the contrastive signal through backpropagation.
The motivation behind the negative signal is to align the embedding of $ A \setminus C $ with that of $ C $, while increasing the distance between the embeddings of $ B \setminus C $ and $ C $, since $ B \setminus C $ leads to an incorrect solution. As noted in \citep{10.5555/3524938.3525859}, contrastive learning has the net effect of pulling positive pairs together while scattering negative examples apart.

\section{Asymptotics of $\mathcal{L}_{\text{contrastive}}$}
According to Theorem 1 in \citep{10.5555/3524938.3525859}, for fixed $ \tau > 0 $, as the number of negative samples $ M \to \infty $, the constrative loss converges to

\begin{align}
& \lim_{M \to \infty} \mathcal{L}_{\text{contrastive}}(f; \tau, M) - \log M \nonumber \\ & = -\frac{1}{\tau} \mathbb{E}_{(x,y) \sim p_{\text{pos}}} [ f(x)^\top f(y) ] \nonumber \\ & + \mathbb{E}_{x \sim p_{\text{data}}} \big[ \log \mathbb{E}_{x^- \sim p_{\text{data}}} [ e^{f(x^-)^\top f(x)/\tau} ] \big].
\end{align}

Hence, the contrastive learning signal can be decomposed into two components: the \textit{alignment term} and the \textit{uniformity term}. The  \textit{alignment term} minimizes the distance between embeddings of positive pairs, while the \textit{uniformity term} encourages negative embeddings to be more uniformly distributed. In \TheName{}, the contrastive learning signal helps align the on-policy sampled CoT distribution with the annotated CoT. This effectively leverages the information contained in the annotated CoT and enhances training stability.

\end{document}